# On the Computational Complexity of Structural Generalization

**Zichao Wei**
Department of Language Science and Technology
Saarland University
Saarbrücken, Germany
ziwe00001@stud.uni-saarland.de

## Abstract

Structural generalization has been measured repeatedly by several benchmarks, yet it has never been formally defined. We give a definition that translates the two premises (compositional structure and unbounded generalization) into mathematical language. The definition itself is neutral: a compiler that hard-codes the rules satisfies it just as well. But structural generalization becomes a scientific question only insofar as the capacity can autonomously emerge from finite data. This question pits the computational lower bound $NC^1$ against the learnable ceiling $TC^0$ of pure Transformers. Under a Montagovian instantiation, each compositional rule splits into two projections: a syntactic face ($F_\gamma$) and a semantic face ($G_\gamma$). Tree evaluation on the $G_\gamma$ side is an instantiation of BFVP, which is $NC^1$-complete (Buss, 1987). A pure Transformer must learn both faces at once, but Kraus et al. (2026) prove that its learnable class $\subseteq TC^0$. Under the standard assumption $TC^0 \neq NC^1$, a pure Transformer cannot learn structural generalization. Neuro-symbolic systems achieve the best benchmark scores precisely because they inject $G_\gamma$, sidestepping the genuinely hard half. Benchmark scores cannot distinguish "learned" from "given." This is what this paper sets out to make clear.

## 1 Introduction

We believe that structural generalization is hard for purely neural networks. Yao and Koller (2022) and SLOG(Li et al., 2023) provide empirical evidence: on categories that require structural transfer, Transformers reach only 40.6%, while neuro-symbolic systems injected with explicit syntactic knowledge reach 70.8%. This evidence is enough to support a belief, but for a proof, an asymmetry remains.

When we voice the belief that "Transformers cannot do structural generalization," we have an intuitive grasp, expressible in ordinary language, of what Transformers and structural generalization mean. But once a reliable proof is demanded, an asymmetry emerges: Transformers have a precise definition, and their computational power is characterized exactly in terms of circuit complexity(Merrill and Sabharwal, 2023), while structural generalization remains unclear. Everything we know about it rests on a handful of benchmarks and a few operational descriptions. COGS describes it as "generalization to novel combinations of familiar primitives and grammatical structures"(Kim and Linzen, 2020); SLOG speaks of "interpret syntactic structures that are themselves unfamiliar from training"(Li et al., 2023); Hupkes et al. (2020) explicitly declines to give a definition. One side of the inequality has mathematics; the other does not. Turning the belief into a theorem requires not more experiments, but a formal definition of structural generalization.

So, what is structural generalization? Literally, it is the generalization of structure. This splits into two parts: "structure" and "generalization." Both terms are used widely across different academic communities, each with its own intension and extension. But we need not debate "structure in general" or "generalization in general" in a vacuum. It is better to return to the community's research practice, uncover the premises it already assumes, and make those premises explicit as a definition.

Existing research carries two implicit premises. First, on structure: benchmarks test generalization over syntactic structure, and syntactic structure is compositional — parts are combined into wholes by rules. Structural generalization is the structural half of compositional generalization; its "structure" denotes compositional structure, not an arbitrary mathematical structure, nor some domain-specific structure resistant to formalization. Second, on generalization: COGS and SLOG split training and test sets by derivation depth, with training covering a finite range of compositional configurations and testing requiring the model to be correct on deeper or unseen configurations. This carries an implicit requirement of unboundedness — generalization to any new compositional configuration.

This paper makes these two implicit premises explicit as a single definition: structural generalization is the ability to correctly evaluate a finite set of compositional rules on any new compositional configuration. The definition is neutral — it does not stipulate where the ability comes from, only what structural generalization is. Its immediate computational consequence: recursive nesting of $n$-ary operations produces tree structure, and unbounded tree evaluation is equivalent to the Boolean Formula Value Problem (BFVP), which is $\mathrm{NC}^1$-complete(Buss, 1987).

But a definition is only a ruler. A compiler satisfies it too. Hard-code the rules directly, and evaluation at any depth becomes trivial. Structural generalization becomes a scientific question only insofar as the capacity can autonomously emerge from finite data. This requirement pits the computational lower bound $\mathrm{NC}^1$ against the learnable ceiling $\mathrm{TC}^0$(Kraus et al., 2026) of pure Transformers. The rest of the paper develops this argument.

## 2 Definitions

### 2.1 Postulates

In the introduction we extracted two premises from existing research. This section makes them explicit as postulates.

The "structure" in existing benchmarks denotes syntactic structure, and syntactic structure is compositional: parts are combined into wholes by rules. This differs from the notion of "structure" used in other disciplines. Partee et al. (1990)'s principle of compositionality attempts to characterize this compositionality, but Zadrozny (1994) proves that the principle is formally vacuous unless a specific algebraic structure is specified: for any language $L$ and any meaning assignment $\mu$, a compositional semantic function satisfying the principle can always be constructed. For "compositionality" to be a substantive notion, an algebraic structure must be specified.

The "generalization" in existing benchmarks also has a specific meaning. COGS(Kim and Linzen, 2020) and SLOG(Li et al., 2023) split training and test sets by derivation depth: training covers a finite range of compositional configurations, and testing requires the model to be correct on deeper or unseen configurations. Hupkes et al. (2020) explicitly declines to give a definition ("we do not aim to provide a new definition of compositionality"), offering instead five behavioral test dimensions. These characterizations are all operational: they specify how to test, not what is being measured. An operational definition conflates the object of measurement with the means of measurement — a test result cannot tell us whether the model learned the rule, or merely learned a substitute whose output happens to agree on the test set.

We make these two implicit premises explicit as two postulates:

> **Postulate 1** (Compositional Structure): the "structure" in structural generalization is compositional — parts are combined into wholes by a finite set of $n$-ary operations.
>
> **Postulate 2** (Unbounded Generalization): the "generalization" in structural generalization is unbounded — it requires correctness on any new compositional configuration, not merely on the finite range covered by training.

### 2.2 Definition

Under Postulates 1 and 2, we give the following formalization.

Let $\Gamma = \{g_1, ..., g_n\}$ be a finite set of operations, where each $g_i$ takes $k_i \geq 1$ inputs and produces one output, with inputs and output drawn from the same finite domain $M$. $\Gamma$ and $M$ together constitute the **means** (Humboldt (1836)). An **expression** $e$ is defined recursively by the following rules:

- (Base) $e = a$, where $a \in M$;
- (Composite) $e = g_{i(e_1, ..., e_{k_i})}$, where $g_i \in \Gamma$ and each $e_j$ is itself an expression.

The **evaluation** $[\![e]\!]$ of an expression $e$ proceeds bottom-up along the nested structure, applying the corresponding operation at each layer:

- (Base) if $e = a$, then $[\![e]\!] = a$;
- (Composite) if $e = g_{i(e_1, ..., e_{k_i})}$, then $[\![e]\!] = g_{i([\![e_1]\!], ..., [\![e_{k_i}]\!])}$.

**Definition 1. (Structural Generalization)** *A system* ***has structural generalization capacity*** *over the means* $(\Gamma, M)$ *if and only if it correctly evaluates every expression* $e$ *generated by* $\Gamma$ *(of any depth and any compositional configuration), i.e.,* $\mathcal{A}(e) = [\![e]\!]$.

When $\Gamma$ contains an operation of arity $k_i \geq 2$, recursive nesting produces branching: each of an operation's multiple inputs can itself be the result of a subexpression. This makes the structure of an expression a tree. Operations of arity 1 (unary transformations) produce only chains, never branches. Compositionality (Postulate 1) is formally the existence of a multi-argument operation, and the tree is its natural graph-theoretic counterpart. Unboundedness (Postulate 2) guarantees that "every expression" is an infinite set; compositional configurations of any depth, whether inside or outside the training distribution, fall under the requirement of Definition 1.

### 2.3 Instantiation in Natural Language: The Montagovian Homomorphism

To apply Definition 1 to natural language, the specific content of $\Gamma$ and $M$ must be given. Montague (1970)'s compositional semantics provides the standard instantiation.

Let $\mathcal{E} = \left(E, (F_\gamma)_{\gamma \in \Gamma}\right)$ be the syntactic algebra and $\mathcal{M} = \left(M, (G_\gamma)_{\gamma \in \Gamma}\right)$ the semantic algebra. A Montagovian semantics is a homomorphism $m : \mathcal{E} \to \mathcal{M}$ satisfying

$$m\big(F_\gamma(e_1, ..., e_k)\big) = G_\gamma(m(e_1), ..., m(e_k)). \quad (1)$$

In natural language, each rule $\gamma \in \Gamma$ in the homomorphism equation has two projections: $F_\gamma$ is the syntactic face (specifying how subexpressions combine structurally), and $G_\gamma$ is the semantic face (specifying how the semantic value is computed once they combine). The two are two faces of the same compositional rule, bound together in the homomorphism equation. The "operation" in Definition 1, under the Montagovian instantiation, comprises both $F_\gamma$ and $G_\gamma$: structural generalization requires not only correct semantic computation ($G_\gamma$) but also correct syntactic composition ($F_\gamma$).

The separation of syntactic and semantic algebras arises naturally in this instantiation. This is a property of natural language, not a requirement of the definition. The syntax/semantics distinction surfaces only when the definition is applied to a natural language with an independent syntactic representation.

The Montagovian homomorphism encodes Humboldt (1836)'s "infinite use of finite means" as a single mathematical object: a finite set of operations together with a lexicon constitutes the finite means; the homomorphism property guarantees that this finite specification extends to expressions of any depth, constituting the infinite use. The same intuition appears under different names in cognitive science. Fodor and Pylyshyn (1988) distinguishes systematicity (recombining roles within the same configuration) from productivity (new expressions of arbitrary complexity). Chomsky (1965)'s competence/performance distinction points to the same core.

## 3 Learnability

Definition 1 is a neutral statement at the level of capacity. A compiler that hard-codes the evaluation rules satisfies Definition 1 perfectly: zero training examples, correct evaluation at any depth. If the entire meaning of structural generalization were "there exists a system with this capacity," the compiler solved it back in the 1960s. Structural generalization is a scientific question not because of the existence of the capacity, but because of its source.

### 3.1 The Triviality of the Compiler

Given the CFG rule set of COGS, a programmer can write the rules directly into a parser. For nested expressions of any depth, the compiler outputs the correct logical form. It passes perfectly under the ruler of Definition 1.

But no one would say the compiler "solved structural generalization." The compiler discovered nothing from data. It only executed rules it was told. Structural generalization is a scientific question not because of "whether it can be done" (the compiler has already done it), but because of "whether it can be done by itself, from finite data."

### 3.2 The Centrality of Learning

The community's own practice has already established learning as the core. COGS and SLOG have a training set and a test set: the training set covers only a finite range of compositional configurations, and the test set requires correctness on unseen configurations. If injecting the rules were legitimate, the training set would be pointless — one could simply write the rules into the system. The existence of a training set indicates that the designers presuppose the capacity must come from data.

This presupposition is continuous with the Humboldtian tradition. In the "infinite use of finite means," the means are formed autonomously from finite experience, not injected from outside. Chomsky's poverty-of-the-stimulus argument points in the same direction: linguistic input underdetermines the grammar, yet the child learns it anyway. The rule emerges from within.

So when the community asks "Can neural models compositionally generalize," the term neural model already smuggles in learnability: something learned from training data by gradient descent. A compiler is not a neural model. A neuro-symbolic system learns only half. A pure Transformer is learning all of it. The three are not facing the same question.

### 3.3 Two Faces, Two Difficulties

Under the Montagovian instantiation, every compositional rule has two projections: $F_\gamma$ (the syntactic face — where which rule applies) and $G_\gamma$ (the semantic face — how the semantic value is computed once the rule is triggered). The two differ essentially in learning difficulty.

On the $F_\gamma$ side: given a linear sequence of words, determine which combines with which. This is a sequence-to-structure mapping. In practice, it can be reduced to supertagging: predicting the syntactic type of each word from its surrounding context. A formal complexity characterization of the $F_\gamma$ side is left to future work.

On the $G_\gamma$ side: given an unknown compositional configuration, compute its semantic value. This requires node-by-node bottom-up evaluation along a tree structure of any depth. This is tree evaluation. The computational complexity of tree evaluation is not covered by constant depth.

Neuro-symbolic systems (represented by AM-Parser(Groschwitz et al., 2018)) inject $G_\gamma$ and learn only $F_\gamma$. Their success confirms exactly this split: learning $F_\gamma$ is feasible; learning $G_\gamma$ is where the real difficulty lies. The next section gives the formal difficulty bound on the $G_\gamma$ side, and why pure Transformers fall short of it.

## 4 The Difficulty of Learning

§3 established that the core of structural generalization is learning, and that the real difficulty of learning lies on the $G_\gamma$ side, i.e., tree evaluation. This section gives the formal difficulty bound.

### 4.1 A Computational Lower Bound

The "evaluation" in Definition 1 has a precise computational complexity. When $\Gamma$ contains an operation of arity $\geq 2$, an expression is a tree structure, and evaluation applies operations node by node, bottom-up along the tree. The complexity of this process depends on the properties of $\Gamma$. We establish a lower bound by constructing a minimal Montagovian semantic fragment.

**Theorem 2. (BFVP)** *(Buss, 1987) The Boolean Formula Value Problem (BFVP) — given a nested AND/OR/NOT formula and a variable assignment, determine its truth value — is* $\mathrm{NC}^1$*-complete (under* $\mathrm{AC}^0$ *reductions).*

We show that there exists a Montagovian semantic fragment whose evaluation encodes BFVP. We choose a Boolean fragment because it is a purely compositional semantic system: finitely many rules, finitely many semantic values, with all complexity arising from nesting depth. This is exactly what structural generalization is meant to capture: the rules are finite and fixed, and only the compositional depth varies.

The formal grammar corresponding to BFVP is:

$$\varphi \to \varphi \wedge \varphi \quad | \quad \varphi \vee \varphi \quad | \quad \neg\varphi \quad | \quad 0 \quad | \quad 1$$

**Proposition 3. (Montagovian Evaluation)** *There exists a Montagovian semantic fragment whose evaluation problem is* $\mathrm{NC}^1$*-hard.*

*Proof sketch.* Construct a "Boolean fragment" grammar $G_B$ of English:

$$S \to \text{both } S \text{ and } S \quad | \quad \text{either } S \text{ or } S \quad | \quad \text{it is not the case that } S \quad | \quad \text{true} \quad | \quad \text{false}$$

$G_B$ is a legitimate Montagovian semantic fragment. Its syntactic algebra is defined by the productions above, its semantic algebra is $\mathcal{M}_B = (\{0,1\}, \{\mathrm{and}, \mathrm{or}, \mathrm{not}\})$, and the homomorphism $m_B$ maps each sentence to its truth value.

There is a direct reduction between $G_B$ and BFVP. Define a translation $\tau$ that replaces $\wedge$ in the formal grammar with "both…and," $\vee$ with "either…or," $\neg$ with "it is not the case that," 0 with "false," and 1 with "true." Then for any Boolean formula $\varphi$, $m_B(\tau(\varphi)) = \mathrm{val}(\varphi)$.

For example, the formula $(1 \wedge (\neg 0)) \vee 0$ translates to "either both true and it is not the case that false or false," whose Montagovian evaluation yields 1, matching the Boolean evaluation of the original formula.

$\tau$ is a symbol-by-symbol local substitution, hence in $\mathrm{AC}^0$. Therefore BFVP $\underset{\mathrm{AC}^0}{\leq}$ evaluation under $m_B$, and Montagovian semantic evaluation is $\mathrm{NC}^1$-hard. □

Proposition 2 establishes the computational lower bound on the $G_\gamma$ side: the complexity of tree evaluation is at least $\mathrm{NC}^1$-complete. This lower bound is not confined to the Boolean fragment $G_B$. Any Montagovian semantic fragment whose grammar contains $G_B$ as a subgrammar has an evaluation problem at least as hard as evaluation under $G_B$ (since the input can always be restricted to the $G_B$ subfragment), i.e., $\mathrm{NC}^1$-hard. Empirically, coordination (and/or), negation (not), and recursive embedding are known universal properties of natural language(Dahl, 1979; Haspelmath, 2007). Therefore, Montagovian semantic evaluation for any natural language we care about is at least $\mathrm{NC}^1$-hard.

### 4.2 The Learnable Ceiling of Pure Transformers

A constant-depth Transformer (without CoT) is $\subseteq \mathrm{TC}^0$(Merrill and Sabharwal, 2023). Under the standard assumption $\mathrm{TC}^0 \subsetneq \mathrm{NC}^1$, an $\mathrm{NC}^1$-hard problem does not belong to $\mathrm{TC}^0$, so a constant-depth Transformer lacks the expressive power tree evaluation requires(Yehudai et al., 2025).

CoT changes the expressive power: a Transformer with CoT and hand-set weights can simulate any Turing machine(Pérez et al., 2019). But expressive power is not learnability. Kraus et al. (2026) prove that, under a finite alphabet and standard positional encoding, the class of functions a CoT Transformer can learn with length generalization remains $\subseteq \mathrm{TC}^0$.

**Theorem 4. (CoT Learnability Ceiling)** *(Kraus et al., 2026, Theorem 3.4) Let* $\mathcal{L} \subset \Sigma^*$*, and let* $F_\mathcal{L}$ *be its indicator function. If* $F_\mathcal{L}$ *has a CoT in C-RASP[Pos], then* $\mathcal{L}$ *is definable in* $\mathrm{TC}^0$*.*

**Theorem 5. (Length Generalization Failure)** *(Kraus et al., 2026, Corollary 3.5) If $\mathcal{L} \notin \mathrm{TC}^0$, then under the learning model of Kraus et al., length generalization fails for infinitely many $n$.*

### 4.3 Impossibility

**Proposition 6. (Impossibility)** *(Structural generalization is not learnable) By Proposition 2, Montagovian semantic evaluation $\notin \mathrm{TC}^0$ (given $\mathrm{TC}^0 \subsetneq \mathrm{NC}^1$). By Theorem 4, this evaluation cannot be learned with length generalization by a finite-alphabet CoT Transformer.*

*Proof sketch.* By Proposition 2, Montagovian semantic evaluation is $\mathrm{NC}^1$-hard. Under $\mathrm{TC}^0 \subsetneq \mathrm{NC}^1$, an $\mathrm{NC}^1$-hard problem does not belong to $\mathrm{TC}^0$. By Theorem 4, a problem outside $\mathrm{TC}^0$ fails length generalization for infinitely many $n$. Growth in the complexity of compositional configurations necessarily entails growth in encoding length, so failure of length generalization entails failure of compositional-configuration generalization. □

Taken together: the $G_\gamma$ side demands at least $\mathrm{NC}^1$. The learnable class of a pure Transformer is $\subseteq \mathrm{TC}^0$. Under standard assumptions the two cannot be equal. The $G_\gamma$ side (tree evaluation) is $\mathrm{NC}^1$-hard. Neuro-symbolic systems inject exactly $G_\gamma$, sidestepping the genuinely hard half and learning only $F_\gamma$. This explains why AM-Parser scores higher than pure Transformers on the benchmark: not because it "learns structural generalization better," but because it never learns the hardest half. The score gap reflects a difference in learning task, not a difference in system capacity.

## 5 Discussion

### 5.1 Why Unboundedness Is Needed

Readers unfamiliar with the compositional generalization literature may find "unbounded" in Postulate 2 puzzling. It does not look like a self-evident assumption on its face — in the typical context of machine learning, generalization usually means training on sequences of length $l$, testing on $2l$, and watching the decay curve. Requiring "any," or unboundedness — correctness on any new compositional configuration whatsoever — sounds excessively demanding.

But this same principle is already practiced in the design of the core benchmarks in the compositional generalization field. SLOG(Li et al., 2023) tests recursion depths 3–12. Recursion depth in natural language rarely exceeds 5(Karlsson, 2007), and humans struggle even with three levels of center embedding. SLOG tests up to 12 not because depth 12 would occur in any realistic distribution (clearly it would not), but because deeper nesting "is still grammatical." In the words of SLOG's own authors: "our goal with SLOG is to evaluate the linguistic competence of NLP models, whose goal is not to simulate human performance limitations." COGS(Kim and Linzen, 2020) is designed by the same logic: training at depths 0–2, testing at 3–12.

Depth 12 and unboundedness are not two different requirements. SLOG is a practical benchmark; it cannot test to infinity, so it truncates at 12. But our formal definition offers no natural reason for a truncation — why stop at 12 and not 13? The rule itself carries no depth limit, so we naturally generalize it to be unbounded. Behind both numbers lies the same spirit: the rule should hold consistently on any compositional configuration not yet seen. Humboldt (1836) calls this "the infinite use of finite means," Chomsky (2000) calls it "discrete infinity"[1], and Fodor and Pylyshyn (1988) frames it as the distinction between productivity and systematicity, the former concerning the extension of a rule to any depth, the latter concerning the recombination of roles at the same depth. What Postulate 2 does is not impose an additional standard stricter than existing benchmarks, but state the design motivation behind those benchmarks in mathematical language.

SLOG's designers chose linguistic competence rather than human performance as the evaluation standard. What the unboundedness postulate does is spell out the mathematical consequence of that "rather than." To reject Postulate 2, one must either give a natural reason for truncating at some particular point, or concede that SLOG's design motivation does not hold up mathematically. If Postulate 2 is accepted, the NC[1] lower bound follows automatically.

[1]Here "infinite" (*unendlich*) should be understood as potential infinity: for any finite compositional configuration, a more complex legitimate configuration can always be constructed, rather than requiring the actual, simultaneous infinite existence of all expressions. Hauser et al. (2002) argues "there is no longest sentence," because any sentence $s$ can be embedded as "Mary thinks that $s$." This is the standard construction of potential infinity.

### 5.2 The Mirage: Why Simple Improvements Do Not Work

The impossibility theorem of §4 means that the function $\hat{m} \in \mathrm{TC}^0$ learned by a CoT Transformer cannot mathematically be the same function as the compositional rule $m \notin \mathrm{TC}^0$ (under the standard assumption $\mathrm{TC}^0 \subsetneq \mathrm{NC}^1$, $\hat{m} = m$ would imply $m \in \mathrm{TC}^0$, a contradiction). $\hat{m}$ can agree with $m$ perfectly on a finite range of compositional configurations, but must diverge on infinitely many more complex ones(Kraus et al., 2026, Corollary 3.5). We therefore call $\hat{m}$ the **mirage** of $m$: indistinguishable from the real thing within a finite observation window, but not the real thing.

The mirage gives a unified explanation for the empirical performance of existing approaches. Donatelli and Koller (2023) observed that T5, after compositional data augmentation, "only learns to correctly parse, for instance, PP recursion up to the depth to which the training data were augmented" — it stops wherever the augmentation stops. Under the mirage framework, what data augmentation does is enlarge the mirage's coverage: augmentation covers more compositional configurations, so the range on which $\hat{m}$ agrees with $m$ grows. But $\hat{m}$ remains a $\mathrm{TC}^0$ function, and it still diverges from $m$ beyond the boundary of augmented coverage. A bigger mirage is still a mirage.

The same logic applies to every improvement that does not change the architecture's computational class. Improvements in training strategy (such as STEP, Lindemann et al. (2024)) change how the search proceeds within the $\mathrm{TC}^0$ function space, not the search space itself. A larger model increases the capacity of the $\mathrm{TC}^0$ lookup table, but the lookup table remains within $\mathrm{TC}^0$. Alternative positional encoding schemes (NoPE, APE, RoPE, AliBi) are all covered by Kraus's theorem. All of these can raise benchmark scores, but the improvement records a better mirage, not a crossing from mirage to the real thing.

The mirage cannot in principle be detected by any finite benchmark. Perfect performance on any finite test set can be achieved by a $\mathrm{TC}^0$ function (since $\mathrm{TC}^0$ can express an arbitrary finite lookup table). Therefore, no matter how high a benchmark score is, it cannot serve as evidence that "the compositional rule was learned."

### 5.3 Iteration and Bandwidth

Unboundedness requires iteration: to evaluate correctly at any depth, the number of computation steps must grow with input depth. Iteration is unavoidable; the question is what form of iteration.

CoT is one form of iteration. A Transformer extends its computation steps by autoregressively generating intermediate tokens, with each step's output becoming the next step's input. But Kraus et al. (2026)'s proof reveals where this bottlenecks: under a finite alphabet $\Xi$, the positional-counting predicates of C-RASP[Pos] can accumulate at most $O(\log N)$ bits of effective information per step(Kraus et al., 2026, Appendix A.2). $O(\log N)$ bits is enough to be hard-coded by a $\mathrm{TC}^0$ circuit, so no matter how long the CoT chain, the learnable function class never exceeds $\mathrm{TC}^0$. The bottleneck is not the iteration itself, but its bandwidth.

Kraus et al. (2026) give positive evidence: when the alphabet size is unbounded (C*-RASP, using signpost tokens), a CoT Transformer reaches Turing completeness (Theorem 3.8). But an unbounded alphabet means the means themselves are no longer finite, incompatible with Postulate 1. To break the bandwidth bottleneck while keeping a finite alphabet, iteration in a continuous latent space (depth-recurrent Transformers, latent reasoning) is one possible route: the input and output still use a finite alphabet, but internal iteration proceeds in a real-valued vector space, and the per-step bandwidth is not constrained by discrete tokens. Such architectures fall outside the scope of Kraus's Theorem 3.4, because the theorem's premises do not apply to continuous hidden states. But removing the bandwidth bottleneck does not mean the problem is solved. Hahn and Rofin (2024) point to a second obstacle: when the iterative module uses global attention, the target function corresponding to the compositional rule has high average sensitivity, the correct solution sits in a sharp minimum of the loss landscape, and SGD is systematically biased toward low-sensitivity shortcuts. Whether latent-space iteration can genuinely break through the $\mathrm{TC}^0$ ceiling remains an open question.

## 5.4 Neuro-Symbolic Systems: Splitting Along $TC^0/NC^1$

Neuro-symbolic systems handle the $NC^1$ threshold differently: rather than trying to get the neural network to cross it, they accept its existence and split the problem along the $TC^0$ / $NC^1$ boundary.

AM-Parser(Groschwitz et al., 2018) represents this approach. A Transformer predicts an AM dependency label for each word: which words are in a compositional relation, and which algebraic operation each relation uses. This is sequence labeling, pattern matching in a local context — learning on the $F_\gamma$ side. A predefined AM algebra engine then recursively evaluates bottom-up along the predicted dependency tree — injection on the $G_\gamma$ side.

In the language of the Montagovian instantiation, this split falls exactly on the $F_\gamma$ / $G_\gamma$ boundary. The Transformer learns an approximation of $F_\gamma$ (where which rule fires); the symbolic system supplies $G_\gamma$ (how to compute once it fires). As analyzed in §3, the $G_\gamma$ side (tree evaluation) is $NC^1$-hard. AM-Parser's success is not because it learns more powerfully, but because it outsources the genuinely hard half to a hand-designed symbolic engine.

This is structurally isomorphic to the compiler: the compiler injects all the rules, AM-Parser injects half the rules. The difference between them is quantitative (the proportion injected), not qualitative. Donatelli and Koller (2023) distinguishes "handwritten" from "learned" grammars, and AM-Parser's algebraic operations belong to the former. What the neural network learns is "where to use which operation" — the operation itself is given by hand ("built in," (Donatelli and Koller, 2023)).

Given this, the meaning of AM-Parser's 70.8% versus the pure Transformer's 40.6% on SLOG needs to be reconsidered. It does not prove that neuro-symbolic systems are "better" at structural generalization. Comparing the two on the same benchmark is like putting a student with half the answers already given against a student sitting a fully closed-book exam. The score difference records a difference in learning task: one has to learn $G_\gamma$, the other does not. The benchmark score cannot distinguish between these two facts.

This does not mean neuro-symbolic systems have no value. Engineering-wise, splitting the problem along $TC^0/NC^1$, so that each component does its own job within its own complexity class, is an effective strategy. It answers a different but practical question: if the semantic face is reliably given, can a neural network learn to apply it at the right place.

Still, one question deserves consideration: to what extent can the semantic face be reliably given? The neuro-symbolic paradigm rests on a belief: since some rules are hard to learn yet are mastered by humans and can be simply injected, injecting what should be injected and learning what should be learned is not merely reasonable but prudent. Yet, at least for natural language, are the "rules" of natural language really explicitly and completely represented at the linguistic level? Notably, COGS and SLOG are built on synthetic language data whose rules are defined a priori. Natural language is different. For example, COGS's obj_pp_to_subj_pp presupposes that PP modification can freely migrate from object position to subject position, but the naturalness of this migration in English depends on the specific verb and its semantics, and is not an unconditional syntactic rule. Beyond English the problem sharpens: SOV languages (such as Japanese, Korean, Turkish) use postpositions rather than prepositions, and modifiers precede their heads, so the structural pattern "PP modifies the noun postpositionally" does not directly apply at all. The test categories of COGS/SLOG carry an implicit English-specific word-order assumption; taking them as a universal standard for "structural generalization" calls for more careful cross-linguistic consideration.

## 5.5 Open Questions

First, can latent-space iteration be proven to overcome the $TC^0$ ceiling? Or does Hahn and Rofin (2024)'s sensitivity obstacle constitute a second barrier as solid as the CoT learnability ceiling?

Second, can Kraus's information-theoretic upper bound and Hahn & Rofin's loss-landscape obstacle be reduced to two consequences of the same architectural property?

Third, a per-category complexity characterization. For each test category of existing benchmarks (COGS, SLOG), does its underlying function belong to $TC^0$ or is it $NC^1$-hard? This determines which categories can be passed by a $\mathrm{TC}^0$ mirage, and which constitute a genuine test of structural generalization capacity.

## 6 Conclusion

This paper gives structural generalization a formal definition, from which the impossibility result for pure Transformers follows. The definition translates the two premises (compositional structure and unbounded generalization) into mathematical language. The definition itself is neutral: a compiler satisfies it too. But structural generalization becomes a scientific question only insofar as the capacity can autonomously emerge from finite data. This question pits the computational lower bound $\mathrm{NC}^1$ against the learnable ceiling $\mathrm{TC}^0$ of pure Transformers.

Under the Montagovian instantiation, a compositional rule splits into two projections, $F_\gamma$ (the syntactic face) and $G_\gamma$ (the semantic face). The $G_\gamma$ side (tree evaluation) is $\mathrm{NC}^1$-hard. A formal characterization of the $F_\gamma$ side is left to future work. Neuro-symbolic systems inject exactly $G_\gamma$, sidestepping the genuinely hard half and learning only $F_\gamma$. A pure Transformer tries to learn both faces at once, but its learnable class is bounded by $\mathrm{TC}^0$ and falls short of $\mathrm{NC}^1$.

The $\mathrm{NC}^1$ threshold is not a dead end. Two paths are already visible. Latent-space iteration may break the bandwidth bottleneck while keeping a finite alphabet, but its feasibility remains to be proven. Neuro-symbolic systems achieve structural generalization in engineering terms through a precise projection split, but their success belongs to "execution given the rule," not "recovery of the rule from data." The two answer different questions. When the community compares their scores on the same benchmark, the ambiguity lies in the fact that the score cannot distinguish between these two questions. This is what this paper sets out to make clear.